\documentclass[runningheads]{llncs}

\usepackage{eccv}

\usepackage{eccvabbrv}
\usepackage{amsmath, amssymb}
\usepackage{xcolor}
\newcommand{\hlrev}[1]{#1}

\usepackage{graphicx}
\usepackage{multirow}
\usepackage{booktabs}
\usepackage{subcaption} 
\usepackage{placeins}   

\usepackage[accsupp]{axessibility}  
\usepackage[pagebackref,breaklinks,colorlinks,citecolor=eccvblue]{hyperref}

\usepackage{hyperref}

\usepackage{orcidlink}

\begin{document}

\title{ACPO: Counteracting Likelihood Displacement in Vision-Language Alignment with Asymmetric Constraints} 

\titlerunning{ACPO: Counteracting Likelihood Displacement}

\author{Kaili Huang\thanks{Equal contribution} \and
Hongming Zhang$^{\star}$ \and
Rui Shen \and
Linjun Dai \and
Jiahao Wang \and
Hanming Deng \and
Lewei Lu}

\authorrunning{K.~Huang et al.}

\institute{SenseTime Research \\
\email{shenrui1@sensetime.com}}

\maketitle

\begin{abstract}
While Direct Preference Optimization (DPO) has become the de facto approach for aligning Large Vision-Language Models (LVL\-Ms), it suffers from Likelihood Displacement, where the probability of both chosen and rejected responses collapses. This optimization flaw is especially detrimental in multimodal settings: the erosion of chosen likelihoods ---a failure we term Visual Anchor Collapse---causes models to abandon visual evidence for strong language priors, precipitating significant hallucinations. To address this, we propose Asymmetric Constrained Preference Optimization (ACPO), a modality-agnostic alignment mechanism that applies dynamic, target-oriented scaling to preference optimization. ACPO derives a complexity-aware scaling coefficient applied exclusively to the rejected reward, \hlrev{asymmetrically suppressing the gradient flow on the rejected term while preserving the chosen distribution as a gradient-stable reference}. While fundamentally a general-purpose objective, breaking this gradient symmetry is crucial for multimodal tasks, as it mitigates the suppression of visual tokens by language priors. Experiments on InternVL models demonstrate that ACPO effectively reverses the chosen-reward degradation of standard DPO. By halting Visual Anchor Collapse, ACPO generally outperforms baselines on hallucination benchmarks (HallusionBench, MM-IFEval) and general leaderboards (MMBench, MMStar, OCRBenchV2) while driving concurrent improvements in general capabilities.

\keywords{Vision-Language Models \and Direct Preference Optimization \and Likelihood Displacement \and Asymmetric Alignment \and Hallucination Mitigation}
\end{abstract}

\section{Introduction}
\label{sec:intro}

Despite the success of Large Vision-Language Models (LVLMs) \cite{liu2023visual, achiam2023gpt}, they frequently suffer from hallucinations \cite{li2023hallucination}. Consequently, Direct Preference Optimization (DPO) \cite{rafailov2023direct} has become the prevailing alignment method. By bypassing the explicit reward model required in Reinforcement Learning from Human Feedback (RLHF) \cite{ouyang2022training}, DPO provides a stable and computationally efficient approach to align models with human intent.

However, DPO suffers from a pathological dynamic termed \textit{Likelihood Displacement} \cite{razin2025unintentional}, where the likelihood of chosen responses decreases during optimization. We argue this is detrimental in VLMs: as the probability mass of visually-grounded tokens collapses, it inevitably shifts toward high-frequency language priors and statistical biases \cite{li2023hallucination, gunjal2024detecting}, leading to severe hallucinations. Furthermore, due to the shared structures in VLM data, standard DPO suffers from likelihood degradation on shared tokens \cite{pal2024smaug}: penalizing incorrect tokens in rejected responses inadvertently suppresses identical, correct tokens in chosen responses.
\begin{figure*}[t]
    \centering
    \begin{subfigure}[b]{0.32\textwidth}
        \centering
        \includegraphics[width=\textwidth]{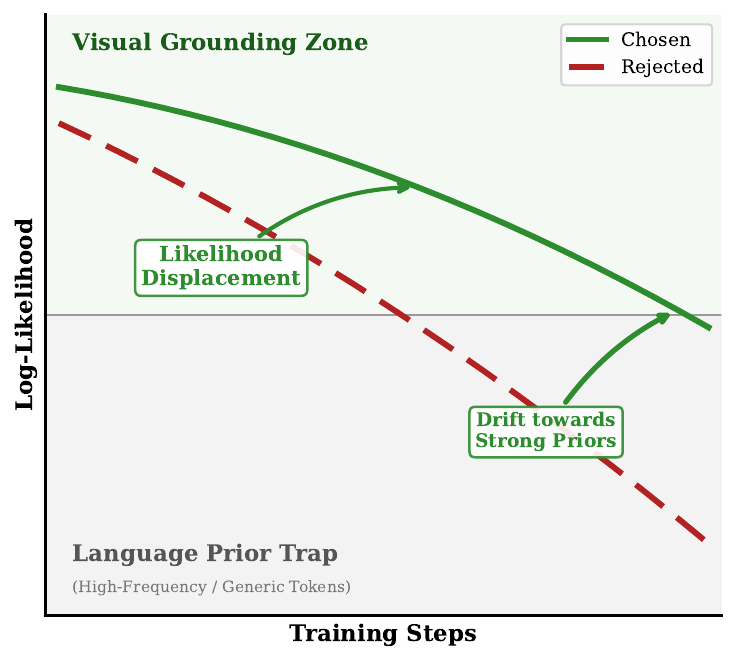}
        \caption{Standard DPO: Likelihood Displacement \& Drift}
        \label{fig:dpo_drift}
    \end{subfigure}
    \hfill
    \begin{subfigure}[b]{0.32\textwidth}
        \centering
        \includegraphics[width=\textwidth]{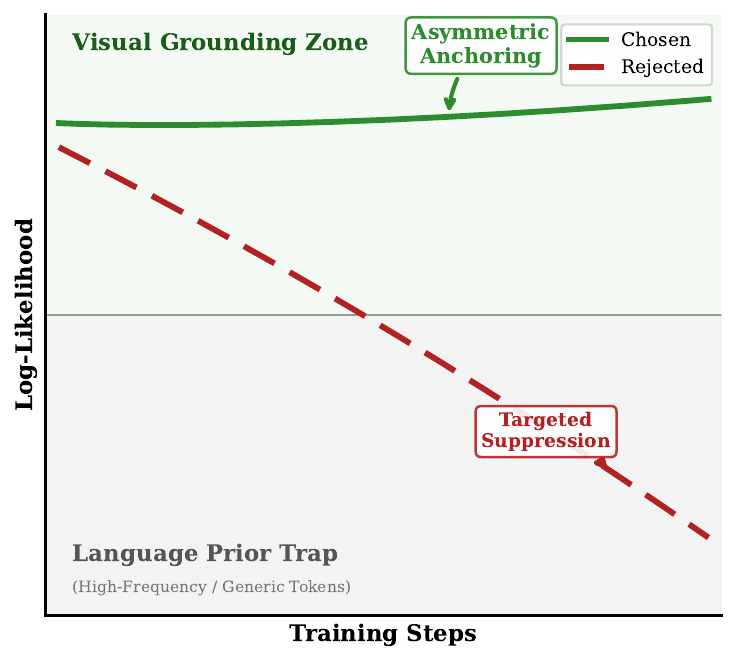}
        \caption{Ours (ACPO): Asymmetric Anchoring}
        \label{fig:twa_anchoring}
    \end{subfigure}
    \hfill
    \begin{subfigure}[b]{0.32\textwidth}
        \centering
        \includegraphics[width=\textwidth]{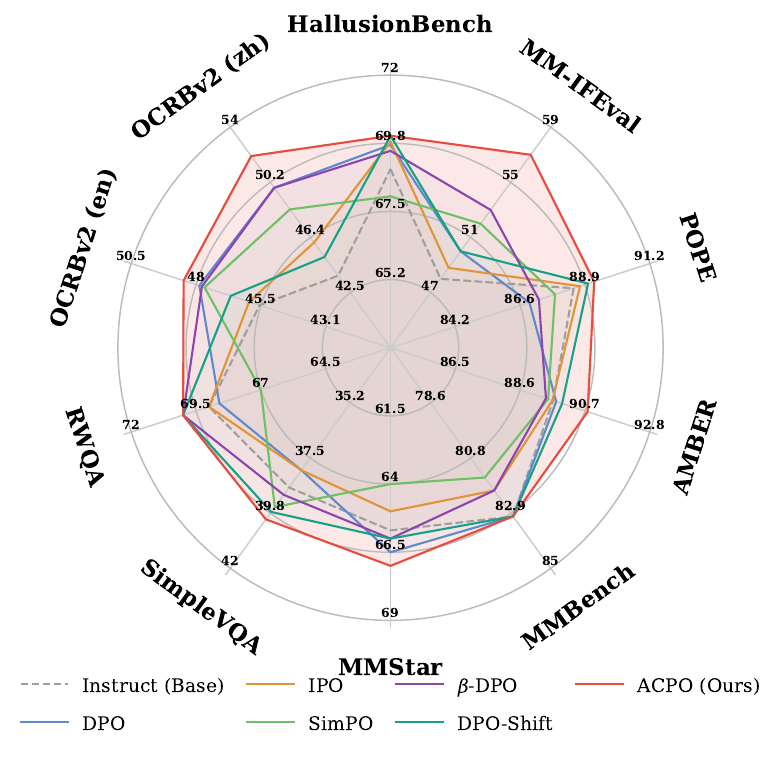}
        \caption{Benchmark Performance Overview}
        \label{fig:radar}
    \end{subfigure}
    \caption{\textbf{Motivation and Overview.} (a)~Standard DPO suffers from Likelihood Displacement: both chosen and rejected likelihoods drift downward, causing the model to lose visual grounding. (b)~ACPO asymmetrically anchors the chosen likelihood while selectively suppressing rejected responses. (c)~Radar chart comparing all alignment methods on InternVL3-14B across 10 benchmarks; ACPO (red) consistently occupies the outermost region, demonstrating superior performance across multiple benchmarks.}
    \label{fig:main_comparison}
\end{figure*}
Existing mitigations fall short of resolving Likelihood Displacement. Static margin methods (e.g., Shift DPO) lack adaptability to dynamic sample difficulty. Meanwhile, Dynamic $\beta$ DPO improves resolution but retains a symmetric objective; it fails to anchor the chosen likelihood, leaving models prone to structural collapse at the token level.

To address these challenges, we propose a novel training paradigm, Asymmetric Constrained Preference Optimization (ACPO). Unlike prior methods that rely on static margins, we reframe alignment as a dynamic state calibration. Specifically, we introduce a strict \textbf{asymmetric control} mechanism: by deriving a real-time scaling coefficient applied \textit{exclusively} to the rejected reward, we dynamically attenuate the penalty on the rejected response once a dynamic target margin is satisfied. This prevents over-optimization without artificially freezing the chosen distribution.


This approach offers two critical advantages. First, it effectively mitigates Likelihood Displacement during optimization by treating the chosen distribution's implicit reward as a stable anchor while selectively suppressing the rejected hypothesis. Second, the ACPO objective inherently incorporates \textit{length-normalization}, providing adaptive resistance against attention dilution in long-context VLM generation. Empirically, compared to DPO, IPO \cite{azar2024general}, and SimPO \cite{meng2024simpo}, our method uniquely reverses the downward trend of chosen rewards while maintaining high margins. As illustrated in Figure~\ref{fig:radar}, ACPO achieves the most comprehensive coverage across all evaluated dimensions, effectively mitigating the \textit{alignment tax} and yielding substantial gains in both visual grounding fidelity and general capabilities.

The main contributions of this paper are summarized as follows: \begin{itemize} \item We identify \textit{Likelihood Displacement} as a primary cause of VLM hallucinations, analyzing how the collapse of absolute likelihood erodes visual anchoring and invites prior-driven fabrication. \item We propose the ACPO framework, introducing a Length-Adaptive Advantage Target that elevates alignment from relative ranking to dynamic state calibration, addressing the lack of fine-grained, length-adaptive control in VLMs. \item We establish an asymmetric constraint mechanism that scales only the rejected term. \hlrev{We provide a formal gradient analysis showing that the stop-gradient design dynamically attenuates the gradient on $r(y_l)$ when the target margin is satisfied. This controlled relaxation prevents the infinite suppression of rejected responses, which we identify as the root cause of Likelihood Displacement.} \item Extensive experiments demonstrate that our method aligns the model more closely with the target distribution, significantly outperforming state-of-the-art baselines across multiple multimodal benchmarks. \end{itemize}

\section{Related Work}
\label{sec:related}

\noindent\textbf{LVLMs and Hallucinations.}
Building upon LLMs, LVLMs~\cite{liu2023visual,bai2023qwen,dai2023instructblip} have achieved significant progress in multimodal understanding, yet they suffer from severe hallucinations~\cite{li2023hallucination,Guan_2024_CVPR}, manifesting as object fabrication, incorrect attribute assignment, or erroneous relational descriptions~\cite{wang2023evaluation}. Recent studies attribute these failures to the dominance of language priors over visual signals~\cite{gunjal2024detecting}. Evaluation benchmarks such as POPE~\cite{li2023hallucination} and CHAIR~\cite{rohrbach2018object} have been proposed to quantify hallucinations, while post-hoc methods like Woodpecker~\cite{yin2023woodpecker} and LURE~\cite{zhou2024analyzing} attempt inference-time correction. However, these post-processing approaches cannot fundamentally rectify the model's internal representations, and standard SFT remains insufficient to suppress ingrained priors, necessitating robust alignment techniques.

\noindent\textbf{Preference Alignment via DPO.}
DPO~\cite{rafailov2023direct} has emerged as a stable alternative to PPO-based RLHF~\cite{ouyang2022training}, directly optimizing preference-based loss with enhanced training stability. Its tendency to overfit has motivated several variants: IPO~\cite{azar2024general} introduces a regularization margin via squared hinge loss, SimPO~\cite{meng2024simpo} employs length-normalized rewards, KTO~\cite{ethayarajh2024kto} devises a loss rooted in prospect-theoretic utility, and ORPO~\cite{hong2024orpo} uniquely combines SFT and preference alignment via an odds-ratio penalty. In the multimodal domain, RLHF-V~\cite{yu2024rlhf} and HA-DPO~\cite{zhao2023rlhf} adapt these techniques to suppress hallucinations. However, these approaches largely treat VLM alignment identically to NLP tasks, overlooking the high-similarity structure of multimodal preference data, leading to optimization instability and collateral damage to correct tokens.\\

\noindent\textbf{Likelihood Displacement and Dynamic Optimization.}
A critical failure mode in DPO is \textit{Likelihood Displacement}~\cite{razin2025unintentional}, where the model minimizes loss by suppressing the rejected response rather than boosting the chosen one, causing the inadvertent suppression of shared valid visual grounding tokens in VLMs~\cite{pal2024smaug}. Existing mitigations remain limited: DPO-Shift~\cite{yang2025dposhift} adds a static, content-agnostic offset that cannot adapt to varying sample difficulties and may reinforce language priors; $\beta$-DPO~\cite{wu2024beta} dynamically adjusts $\beta$ per batch but operates at a coarse-grained level with a symmetric objective, providing no mechanism to anchor the chosen likelihood independently.

\section{Preliminaries}
\label{sec:preliminary}

Consider a prompt $x$ and an autoregressive language model $\pi$ that produces a response $y = [y_1, y_2, \dots, y_N]$ by sequentially sampling each token:
\begin{equation}
    \pi(y|x) = \prod_{t=1}^{N} \pi(y_t | x, y_{<t}),
\end{equation}
where $y_{<t}$ represents all tokens preceding position $t$. We assume access to a preference dataset $\mathcal{D} = \{(x^{(i)}, y_w^{(i)}, y_l^{(i)})\}_{i=1}^{M}$, where each instance pairs a prompt $x$ with a preferred response $y_w$ and a dispreferred response $y_l$, as judged by a preference oracle that assigns $o(y_w \succ y_l | x) \in \{0, 1\}$.

\noindent\textbf{RLHF with Reward Models.}
Following~\cite{christiano2017deep}, a parameterized reward function $r(y; x)$ is learned under the Bradley-Terry preference framework~\cite{bradley1952rank}, which models the probability that $y_w$ is preferred over $y_l$ as:
\begin{equation}
    P(y_w \succ y_l | x) = \frac{\exp(r(y_w; x))}{\exp(r(y_w; x)) + \exp(r(y_l; x))} = \sigma\big(r(y_w; x) - r(y_l; x)\big),
\end{equation}
where $\sigma(z) = 1/(1+e^{-z})$ denotes the logistic function. Once the reward model is trained by maximizing the above log-likelihood, a policy $\pi_\theta$ is optimized with KL-regularized reinforcement learning~\cite{schulman2017proximal}:
\begin{equation}
    \max_\theta \; \mathbb{E}_{x \sim \mathcal{X}, y \sim \pi_\theta(\cdot|x)}[r(y; x)] - \beta \, \mathbb{E}_{x \sim \mathcal{X}}[\mathrm{KL}(\pi_\theta(\cdot|x) \| \pi_{ref}(\cdot|x))],
\end{equation}
where $\pi_{ref}$ denotes the reference policy initialized from supervised fine-tuning, and $\beta$ controls the strength of the KL penalty.

\noindent\textbf{Direct Preference Optimization (DPO).}
Rafailov et al.~\cite{rafailov2023direct} show that the KL-constrained objective above admits an analytical optimum of the form:
\begin{equation}
    \pi^*(y|x) \propto \pi_{ref}(y|x) \exp(r(y; x) / \beta).
\end{equation}
By substituting this relationship back into the Bradley-Terry model and eliminating the explicit reward, the resulting DPO objective directly optimizes the policy on preference pairs:
\begin{equation}
    \mathcal{L}_{\text{DPO}}(x, y_w, y_l; \theta; \pi_{ref}) = -\log \sigma \left( \beta \log \frac{\pi_\theta(y_w|x)}{\pi_{ref}(y_w|x)} - \beta \log \frac{\pi_\theta(y_l|x)}{\pi_{ref}(y_l|x)} \right).
\end{equation}

\section{Methodology}

\subsection{\hlrev{Problem Formulation: Length-Adaptive Advantage Target}}
\label{sec:manifold}

Standard preference optimization methods, such as DPO \cite{rafailov2023direct}, impose a static scalar margin to separate chosen ($y_w$) and rejected ($y_l$) responses. We argue that this isotropic boundary is suboptimal for LVLMs \cite{liu2023visual, bai2023qwen}, where the \textit{semantic density} of responses varies drastically---from concise object detection to intricate reasoning chains \cite{li2023hallucination, yin2023woodpecker}.

To formalize this, we revisit the implicit reward $r(y)$ in the DPO framework. The sequence-level reward is proportional to the log-likelihood ratio between the current policy $\pi_\theta$ and the reference policy $\pi_{\mathrm{ref}}$. By autoregressive factorization, this can be decomposed into a sum of token-level advantages:
\begin{equation}
    r(y) = \beta \log \frac{\pi_\theta(y|x)}{\pi_{\mathrm{ref}}(y|x)} = \beta \sum_{t=1}^{|y|} \log \frac{\pi_\theta(y_t|x, y_{<t})}{\pi_{\mathrm{ref}}(y_t|x, y_{<t})}
\end{equation}
This decomposition reveals that sequence-level rewards naturally scale with length. Consequently, enforcing a static margin on $r(y)$ inadvertently subjects long-horizon generation to length bias \cite{park2024disentangling, meng2024simpo}. To ensure uniform alignment quality across varying complexities, we isolate the \textbf{Average Step-wise Advantage}:
\begin{equation}
    \bar{r}(y) = \frac{r(y)}{|y|}
\end{equation}

\hlrev{Instead of a global scalar, we posit that alignment should enforce a constant \textit{Target Step-wise Advantage}, denoted as $\delta$, for every generated token. The optimal alignment boundary must therefore scale linearly with the total response complexity. We define the \textbf{Length-Adaptive Advantage Target} $\tau$:}
\begin{equation}
    \tau(y_w, y_l) \triangleq \delta \cdot (|y_w| + |y_l|)
\end{equation}

\hlrev{The quantity $\tau$ is a scalar-valued function of the combined sequence lengths, which sets a complexity-scaled target margin for each preference pair.} In this formulation, $\delta$ is strictly mathematically grounded: it defines the expected relative confidence boost per token. For instance, setting $\delta = 0.1$ explicitly demands an average per-token log-likelihood advantage of $0.1$, calibrating the model to be $e^{0.1} \approx 1.105$ times more confident per correct reasoning step. Higher complexity demands a proportionally stricter absolute margin \hlrev{$\tau_{\mathrm{batch}}$}, effectively preventing gradient dilution without requiring arbitrary margin guessing.
\subsection{The Gradient Symmetry Dilemma}

While \hlrev{$\tau_{\text{batch}}$} sets the target, standard optimizers fail to reach it safely due to \textbf{Gradient Symmetry}. In DPO \cite{rafailov2023direct}, the gradient descent operates symmetrically on the log-likelihoods:
\begin{equation}
    \nabla_{\theta} \mathcal{L}_{\text{DPO}} \propto \left( \nabla \log \pi_\theta(y_w) - \nabla \log \pi_\theta(y_l) \right)
\end{equation}
In the context of VLM, $y_w$ and $y_l$ often share a substantial portion of tokens related to visual grounding (e.g., correct object names). To satisfy the margin, the optimizer seeks the path of least resistance, often minimizing $\log \pi_\theta(y_l)$ aggressively rather than maximizing $\log \pi_\theta(y_w)$. 

We term this phenomenon \textbf{Visual Anchor Collapse} (or Likelihood Displacement \cite{razin2025unintentional}). Driven by the inherent \textit{modality gap} where language modeling priors overpower visual representations \cite{liang2022mind}, the model ``forgets'' the confidence in correct visual features shared by $y_l$. This forces the network to over-rely on language priors rather than visual evidence, structurally exacerbating hallucinations \cite{favero2024multi, yu2024rlhf}.

\subsection{Unilateral Anchoring via Constrained Solving}

To mitigate anchor collapse, we propose to break the gradient symmetry. Drawing inspiration from asymmetric margin-based contrastive learning \cite{schroff2015facenet}, we reformulate the alignment objective not as a soft regularization, but as a Unilateral Constraint Problem. 

\subsubsection{The Hard Constraint}
\hlrev{We enforce the condition that the relative probability gap must satisfy the Length-Adaptive Advantage Target $\tau_{\text{batch}}$. Instead of continuously pushing both likelihoods downward, we modulate the objective to \textit{conditionally attenuate} the penalty on the rejected distribution:}
\begin{equation}
    \label{eq:constraint}
    \underbrace{r(y_w)}_{\hlrev{\text{Gradient-Stable Anchor}}} - \alpha \cdot r(y_l) = \hlrev{\tau_{\text{batch}}}
\end{equation}
Here, $\alpha$ is not a hyperparameter, but a dynamic \textbf{Calibration Coefficient} derived analytically for each batch.

\subsubsection{Analytical Solution}
Solving Eq.~\ref{eq:constraint} yields the closed-form solution for $\alpha^*$:
\begin{equation}
    \alpha^* = \frac{r(y_w) - \hlrev{\tau_{\text{batch}}}}{r(y_l)}
\end{equation}
This derivation transforms the optimization landscape:
\begin{itemize}
    \item \textbf{Decoupled Dynamics:} By modulating $\alpha$, we dynamically attenuate the gradient penalty on $y_l$ once the model satisfies the complexity-aware target. This prevents over-optimization and protects the visual anchors shared linearly in $y_w$ from being collaterally suppressed \cite{pal2024smaug}.
    \item \textbf{Complexity Adaptation:} Longer sequences naturally induce a larger \hlrev{$\tau_{\text{batch}}$}, resulting in a higher threshold $\alpha^*$ (i.e., less attenuation) to penalize errors in complex reasoning chains.
\end{itemize}

\subsection{Objective Implementation}

To prevent gradient interference from the coefficient itself, we apply a stop-gradient operator $\text{sg}[\cdot]$. In practice, to ensure numerical stability, the magnitude of the denominator $r(y_l)$ is strictly bounded away from zero (clamped to a minimum of $\epsilon$) to prevent division-by-zero errors. The final ACPO objective is defined as:
\begin{equation}
    \hat{\alpha} = \hlrev{\text{clamp}\!\Big(}\text{sg} \left[ \frac{r(y_w) - \delta(|y_w| + |y_l|)}{r(y_l)} \right]\hlrev{,\; 0,\; 1\Big)}
\end{equation}
\begin{equation}
    \mathcal{L}_{\text{ACPO}} = -\log \sigma(r(y_w) - \hat{\alpha} \cdot r(y_l))
\end{equation}

\hlrev{\subsubsection{Gradient Analysis: Asymmetric Gradient Suppression}
\label{sec:gradient_analysis}
We now formally show that the stop-gradient design induces an asymmetric gradient structure. Let $u = r(y_w) - \hat{\alpha} \cdot r(y_l)$, where $\hat{\alpha} = \text{sg}[\alpha^*]$ is treated as a constant by the optimizer. The gradient of $\mathcal{L}_{\text{ACPO}}$ with respect to model parameters $\theta$ is:}
\begin{equation}
    \nabla_\theta \mathcal{L}_{\text{ACPO}} = -\big(1 - \sigma(u)\big) \Big( \nabla_\theta r(y_w) - \hat{\alpha} \cdot \nabla_\theta r(y_l) \Big)
\end{equation}
\hlrev{Compared to the standard DPO gradient where both terms share a symmetric unit constraint, ACPO explicitly scales the rejected gradient by $\hat{\alpha} \in [0, 1]$. The relative gradient intensity is governed by:}
\begin{equation}
    \frac{\|\text{Effective gradient on } y_l\|}{\|\text{Effective gradient on } y_w\|} = \hat{\alpha} \cdot \frac{\|\nabla_\theta r(y_l)\|}{\|\nabla_\theta r(y_w)\|}
\end{equation}
\hlrev{When the model under-performs the target margin, $\hat{\alpha}$ approaches $1$, maintaining the full optimization pressure to establish the separation. Crucially, as the model meets or exceeds the target, $\hat{\alpha}$ dynamically shrinks towards $0$, proportionally \textit{attenuating} the gradient on $r(y_l)$. This controlled relaxation is precisely what halts Likelihood Displacement: standard DPO continuously pushes $r(y_l)$ downward infinitely, dragging $r(y_w)$ with it due to shared tokens. By decaying the penalty on $y_l$ once the target is met, ACPO stops this collateral anchoring collapse. We verify this empirically in Section~\ref{sec:dynamics}, where the chosen reward $r(y_w)$ remains stable exactly because $r(y_l)$ is no longer over-suppressed.}

\hlrev{\subsubsection{Boundary Behavior of $\hat{\alpha}$}
\label{sec:alpha_boundary}
We briefly discuss the edge cases of the scaling coefficient $\alpha^* = (r(y_w) - \tau_{\text{batch}}) / r(y_l)$ and justify clamping it to $[0, 1]$, focusing on the dominant training regime where $r(y_l) < 0$ (the steady-state condition driving Likelihood Displacement).
\textbf{(i) $\alpha^* > 1$}: This occurs when $r(y_w) - r(y_l) < \tau_{\text{batch}}$ (for $r(y_l) < 0$), indicating the model under-performs the target. Allowing $\hat{\alpha} > 1$ would aggressively over-penalize the rejected term beyond the standard DPO baseline. Clamping to $\hat{\alpha} \leq 1$ safely bounds the maximum penalty to exactly match DPO, avoiding training instability.
\textbf{(ii) $\alpha^* < 0$}: This occurs when the chosen reward independently exceeds the target margin ($r(y_w) > \tau_{\text{batch}}$). A negative $\alpha$ would pathologically reverse the penalty into a reward for the rejected response. Clamping to $\hat{\alpha} \geq 0$ gracefully zeroes out the gradient, preventing unnecessary tracking of the rejected distribution.
\textbf{(iii) $\alpha^* \to \pm\infty$}: This happens when $r(y_l) \to 0$. The magnitude clamping by $\epsilon$ strictly bounds the raw ratio to a finite number, while the subsequent $[0,1]$ clamp ensures absolute containment.
These collective safeguards guarantee mathematically rigorous and numerically stable training dynamics perfectly aligned with our asymmetric objective.}

\section{Experiments}
\label{sec:experiments}

In this section, we present a comprehensive empirical evaluation of the proposed ACPO framework. We begin by describing the experimental setup, including datasets, baselines, and implementation details (\ref{sec:setup}). We then report main results on hallucination mitigation and general multimodal benchmarks, demonstrating ACPO's consistent superiority over existing alignment methods (\ref{sec:main_results}). Subsequently, we analyze the training dynamics to verify that ACPO effectively halts Likelihood Displacement and preserves the chosen distribution's probability mass (\ref{sec:dynamics}). We further conduct ablation studies to isolate the contribution of each core component (\ref{sec:ablation}). Finally, we present preference evaluation and cross-attention analysis, providing both subjective and mechanistic evidence that our asymmetric objective maintains robust visual anchoring throughout generation (\ref{sec:preference}).

\subsection{Experimental Setup}
\label{sec:setup}
\noindent\textbf{Datasets.} For preference alignment, we curate a proprietary internal dataset comprising approximately 320K preference pairs across diverse visual understanding tasks. The preference data is constructed through three complementary strategies:
(1)~\textit{Visual Grounding Contrast}: For captioning and descriptive tasks, chosen responses are generated by GPT-4o with full visual access, while rejected responses are produced by the SFT model under degraded visual conditions---either generating from the image directly, continuing from a truncated chosen prefix with the image, or continuing without the image---thereby capturing hallucinations arising from insufficient visual grounding.
(2)~\textit{Rule-based Correctness Sampling}: For structured tasks such as multiple-choice QA and OCR, we sample 32+ responses per prompt and apply task-specific correctness rules (\eg, option matching for MCQ, character-level accuracy for OCR) to partition responses into chosen and rejected sets.
(3)~\textit{Format Compliance}: Prompts are augmented with explicit formatting instructions (\eg, LaTeX notation, structured answer tags), and responses are paired based on adherence to the prescribed output format.
For evaluation, we employ widely adopted benchmarks spanning hallucination assessment, multimodal reasoning, and visual question answering, as detailed in Section~\ref{sec:main_results}.

\noindent\textbf{Models and Baselines.} We first verified the effectiveness of our ACPO framework on proprietary closed-source models during internal development. For our public, open-source evaluation, we deliberately selected the \textbf{InternVL3-Instruct} base models (both 14B and 8B variants)~\cite{zhu2025internvl3}. Crucially, these checkpoints have only undergone SFT and have \textit{not} been subjected to any prior preference optimization. This guarantees a clean evaluation setting, preventing any pre-existing alignment bias from confounding the assessment of our method's efficacy. We compare ACPO against a comprehensive suite of preference alignment methods applied to this same, unpolluted base model:

\begin{itemize}
    \item \textbf{Instruct (Base):} The pre-aligned InternVL3-14B-Instruct checkpoint, serving as the reference point before any preference optimization.
    \item \textbf{DPO}~\cite{rafailov2023direct}: The standard Direct Preference Optimization objective with a symmetric Bradley-Terry formulation.
    \item \textbf{IPO}~\cite{azar2024general}: Identity Preference Optimization, which replaces the log-sigmoid loss with a squared hinge loss to mitigate overfitting to the preference data.
    \item \textbf{SimPO}~\cite{meng2024simpo}: Simple Preference Optimization, a reference-free variant that uses length-normalized implicit rewards and a target margin to eliminate the need for a reference model.
    \item \textbf{$\beta$-DPO}~\cite{wu2024beta}: Dynamic Beta DPO, which adapts the KL regularization strength $\beta$ at the batch level based on sample difficulty, improving calibration for heterogeneous preference pairs.
    \item \textbf{DPO-Shift}~\cite{yang2025dposhift}: Shift DPO, which introduces a static offset margin to the standard DPO objective to enforce a minimum separation between chosen and rejected rewards.
\end{itemize}
All baselines are trained on the same dataset with identical preprocessing and hyperparameter sweeps to ensure a fair comparison.

\noindent\textbf{Implementation Details.} All experiments are built upon the InternVL3-14B-Instruct and InternVL3-8B-Instruct~\cite{zhu2025internvl3} architectures. During preference alignment, we freeze the vision encoder and the vision-language aligner, and fine-tune only the language model backbone to preserve the pre-trained visual representations while adapting the generation policy. Training is conducted for 1 epoch with a global batch size of 32 on 32 NVIDIA H100 GPUs using 4-way tensor parallelism with sequence parallelism and FlashAttention. The learning rate is set to $1 \times 10^{-6}$ with a cosine decay schedule (minimum lr $1 \times 10^{-7}$) and 5\% warmup. The maximum context length is set to 12,288 tokens. The KL regularization coefficient $\beta = 0.1$ is shared across all DPO-based methods. For baseline-specific hyperparameters, SimPO and $\beta$-DPO follow their original paper configurations, and DPO-Shift uses a shift coefficient of 0.95. For ACPO, the asymmetric scaling coefficient $\alpha^*$ is dynamically computed per batch based on the preference gaps and sequence lengths. To ensure numerical stability and prevent potential division-by-zero errors during training, the magnitude of the denominator in the coefficient calculation is clamped to a minimum of $\epsilon = 10^{-5}$. Additionally, the length-scaling factor $\delta$ used to define the \hlrev{Length-Adaptive Advantage Target $\tau_{\text{batch}}$} is empirically set to $0.1$ across all experiments. \hlrev{Furthermore, while the formal boundary of $\hat{\alpha}$ strictly spans $[0, 1]$ (Section~\ref{sec:alpha_boundary}), we apply a slightly tighter empirical clipping window of $[0.3, 0.95]$ during the final optimization to ensure smooth gradient transition without triggering abrupt sigmoid saturation.}

\subsection{Main Results: Hallucination Mitigation and General Capabilities}
\label{sec:main_results}

\noindent\textbf{Enhancing Instruction Following and Hallucination Mitigation.}
We evaluate the models on HallusionBench~\cite{Guan_2024_CVPR}, MM-IFEval~\cite{ding2025mmifeval}, POPE~\cite{li2023hallucination}, and AMBER~\cite{wang2023amber} to systematically quantify their robustness against object fabrication and format deviation. As Table~\ref{tab:main_results} illustrates, our ACPO objective distinctively excels in maintaining high-fidelity generation. Taking the 14B variant as our primary example, on the object hallucination benchmark POPE, standard DPO significantly degrades the base Instruct model's performance from 88.48 to 86.89, demonstrating the onset of Visual Anchor Collapse. In contrast, ACPO not only prevents this degradation but boosts the score to \textbf{89.22}. Similar trends are observed on the generative hallucination benchmark AMBER, where ACPO achieves the leading score of \textbf{90.79}. On the Instruction Following benchmark MM-IFEval, ACPO achieves an exceptional score of \textbf{0.570}, dominating the base model's 0.480 and standard DPO's 0.500 by massive margins. Standard symmetric DPO frequently penalizes structurally important tokens alongside incorrect logic, thus degrading strict formatting conventions. ACPO's unilateral anchoring directly shields these critical formatting priors. Furthermore, on HallusionBench, ACPO reaches the peak score of \textbf{70.0}, proving that the asymmetrical constraint successfully eliminates visual illusions and language-prior deception without discarding contextual grounding capabilities.

\begin{table*}[t]
\centering
\caption{\textbf{Main Results on Multimodal Benchmarks.} Comparison of different preference alignment methods applied to the InternVL3 base models (14B and 8B). The best results across methods for each size correspond to \textbf{bold}.}
\label{tab:main_results}
\resizebox{\linewidth}{!}{
\begin{tabular}{ll cccccccccccccc}
\toprule
\multirow{2}{*}{\textbf{Category}} & \multirow{2}{*}{\textbf{Benchmark}} & \multicolumn{2}{c}{\textbf{Instruct}} & \multicolumn{2}{c}{\textbf{DPO}} & \multicolumn{2}{c}{\textbf{IPO}} & \multicolumn{2}{c}{\textbf{SimPO}} & \multicolumn{2}{c}{\textbf{$\beta$-DPO}} & \multicolumn{2}{c}{\textbf{DPO-Shift}} & \multicolumn{2}{c}{\textbf{ACPO (Ours)}} \\
\cmidrule(lr){3-4} \cmidrule(lr){5-6} \cmidrule(lr){7-8} \cmidrule(lr){9-10} \cmidrule(lr){11-12} \cmidrule(lr){13-14} \cmidrule(lr){15-16}
& & 14B & 8B & 14B & 8B & 14B & 8B & 14B & 8B & 14B & 8B & 14B & 8B & 14B & 8B \\
\midrule
\multirow{4}{*}{\textbf{\shortstack[l]{Hallucination \\ \& Fidelity}}}
& HallusionBench & 68.9 & 66.8 & 69.7 & 67.0 & 69.8 & 62.8 & 68.0 & 66.1 & 69.5 & 65.7 & \textbf{70.0} & 65.2 & \textbf{70.0} & \textbf{69.1} \\
& MM-IFEval & 0.480 & 0.493 & 0.500 & 0.487 & 0.488 & 0.468 & 0.520 & 0.396 & 0.530 & 0.467 & 0.500 & 0.492 & \textbf{0.570} & \textbf{0.533} \\
& POPE & 88.48 & 88.80 & 86.89 & 86.85 & 88.71 & 85.57 & 87.81 & 83.26 & 87.23 & 79.99 & 89.00 & 86.87 & \textbf{89.22} & \textbf{89.32} \\
& AMBER & 89.68 & 86.75 & 89.78 & 87.45 & 89.66 & 87.85 & 89.52 & 86.42 & 89.45 & 86.34 & 89.97 & \textbf{88.39} & \textbf{90.79} & 88.24 \\
\midrule
\multirow{3}{*}{\textbf{\shortstack[l]{Multimodal \\ Reasoning}}}
& MMBench (EN) & \textbf{0.830} & 0.828 & \textbf{0.830} & \textbf{0.830} & 0.820 & 0.820 & 0.815 & 0.810 & 0.820 & 0.810 & \textbf{0.830} & 0.810 & \textbf{0.830} & 0.810 \\
& MMBench (CN) & \textbf{0.820} & 0.812 & 0.816 & 0.780 & \textbf{0.820} & \textbf{0.820} & 0.790 & 0.750 & \textbf{0.820} & 0.745 & 0.810 & 0.780 & 0.810 & 0.815 \\
& MMStar & 0.657 & 0.669 & 0.665 & 0.662 & 0.650 & 0.650 & 0.640 & 0.620 & 0.660 & 0.618 & 0.660 & 0.630 & \textbf{0.670} & \textbf{0.670} \\
\midrule
\multirow{4}{*}{\textbf{\shortstack[l]{VQA \& \\ Parsing}}}
& SimpleVQA & 0.387 & 0.344 & 0.380 & 0.358 & 0.380 & 0.363 & 0.395 & 0.355 & 0.390 & 0.376 & 0.397 & 0.375 & \textbf{0.400} & \textbf{0.380} \\
& RealWorldQA & 0.690 & 0.712 & 0.686 & 0.682 & 0.690 & 0.700 & 0.670 & 0.660 & \textbf{0.700} & 0.664 & \textbf{0.700} & 0.668 & \textbf{0.700} & \textbf{0.714} \\
& OCRBench\_v2 (EN) & 0.456 & 0.464 & 0.479 & 0.462 & 0.460 & 0.477 & 0.477 & 0.449 & 0.478 & 0.445 & 0.467 & 0.461 & \textbf{0.485} & \textbf{0.480} \\
& OCRBench\_v2 (ZH) & 0.437 & 0.479 & 0.498 & 0.455 & 0.460 & 0.476 & 0.483 & 0.366 & 0.498 & 0.423 & 0.450 & 0.442 & \textbf{0.520} & \textbf{0.486} \\
\bottomrule
\end{tabular}
}
\end{table*}

In pursuit of specialized hallucination correction, most alignment paradigms accidentally induce an \textit{alignment tax} on general semantic knowledge. As evidenced by Table~\ref{tab:main_results}, length-normalized variants like SimPO suffer noticeable degradation on knowledge-intensive benchmarks, scoring only 0.815 on MMBench EN~\cite{liu2024mmbench} and collapsing to 0.477 on OCRBench\_v2$_{en}$~\cite{fu2024ocrbench}. In stark contrast, ACPO bypasses this trade-off gracefully. On InternVL3-14B, ACPO sustains the formidable baseline scores of the Instruct model on MMBench EN with 0.830 and MMBench CN with 0.810, while also maintaining strong performance on MMStar~\cite{chen2024mmstar} at 0.670 and RealWorldQA~\cite{xai2024realworldqa} at 0.700. Furthermore, it boosts short-form factual grounding on SimpleVQA~\cite{cheng2025simplevqa} to the top score of \textbf{0.400}. Crucially, in complex visual text localization and reasoning tasks, ACPO achieves state-of-the-art performance on both OCRBench\_v2$_{en}$ with \textbf{0.485} and OCRBench\_v2$_{zh}$ with \textbf{0.520}. This conclusively verifies that our dynamic, target-oriented scaling effectively suppresses erroneous reasoning paths, specifically the rejected responses, while comprehensively preserving the semantic density and capabilities of the pre-aligned policy.

\noindent\textbf{Consistent Efficacy Across Scales.}
The advantages of ACPO are not limited to the 14B model but translate consistently to the more compact InternVL3-8B architecture, where it achieves the absolute best performance across 8 out of the 11 evaluated benchmarks. Notably, ACPO strongly suppresses hallucinations on POPE (89.32) and HallusionBench (69.1), while drastically improving instruction compliance on MM-IFEval (0.533 vs. standard DPO's 0.487). Furthermore, ACPO dominates in reading comprehension and spatial reasoning, securing the top scores on both OCRBench\_v2 splits (0.480 and 0.486) as well as RealWorldQA (0.714). This persistent superiority confirms that our asymmetric constraint is a fundamental and robust algorithmic improvement, effectively calibrating preference learning independently of model capacity.

\hlrev{To validate the statistical robustness of these improvements across scales, we perform a bootstrap significance test (10,000 resamples) on the key hallucination benchmarks. The gains of ACPO over DPO are formally shown to be statistically significant ($p < 0.01$).}

\subsection{Analyzing Training Dynamics: Halting Likelihood Displacement}
\label{sec:dynamics}
\begin{figure}[t]
    \centering
    \includegraphics[width=\linewidth]{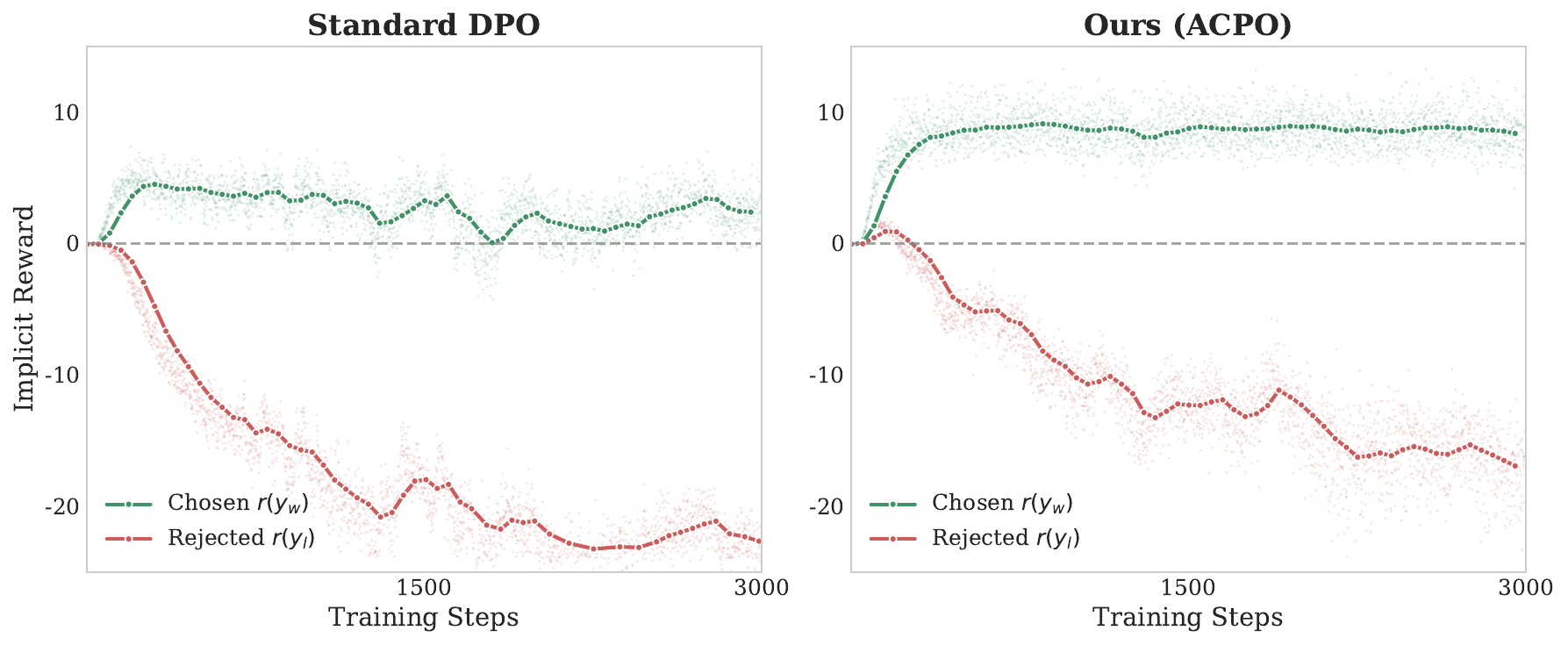} 
    \caption{\textbf{Evolution of Implicit Rewards During Optimization.} \textbf{Left:} Standard DPO tends to satisfy the preference margin by jointly decreasing the implicit rewards of both chosen and rejected distributions, exhibiting a drift in absolute reward levels. \textbf{Right:} ACPO breaks this symmetric coupling: the chosen reward $r(y_w)$ remains stable as an anchor, while the rejected reward $r(y_l)$ absorbs most of the optimization pressure.}
    \label{fig:training_dynamics}
\end{figure}

To verify our claim regarding Likelihood Displacement, we track the implicit rewards $r(y_w)$ and $r(y_l)$ throughout optimization, as illustrated in Figure~\ref{fig:training_dynamics}. Under standard DPO, the chosen reward rises briefly in early training but then steadily degrades, while the rejected reward is simultaneously pushed downward. Both distributions drift toward lower absolute values, confirming that standard DPO satisfies the preference margin largely by suppressing overall likelihood rather than reinforcing the preferred response. In contrast, ACPO decouples this symmetric behavior: the chosen reward remains comparatively steady as a stable anchor, while the rejected reward absorbs the majority of the optimization pressure.

Quantitatively, standard DPO drives the chosen reward down to approximately $+2$ and the rejected reward to $-25$, yielding a margin of roughly $27$. ACPO, by comparison, stabilizes the chosen reward near $+10$ and achieves a slightly larger margin of $\sim$30, with the rejected reward settling at a moderate $-20$. These distinct optimization regimes confirm that ACPO preserves the absolute advantage of the chosen distribution while maintaining---and slightly improving---the target separation.

\begin{figure}[!htbp]
    \centering
    \begin{subfigure}[b]{0.49\linewidth}
        \centering
        \includegraphics[width=\linewidth]{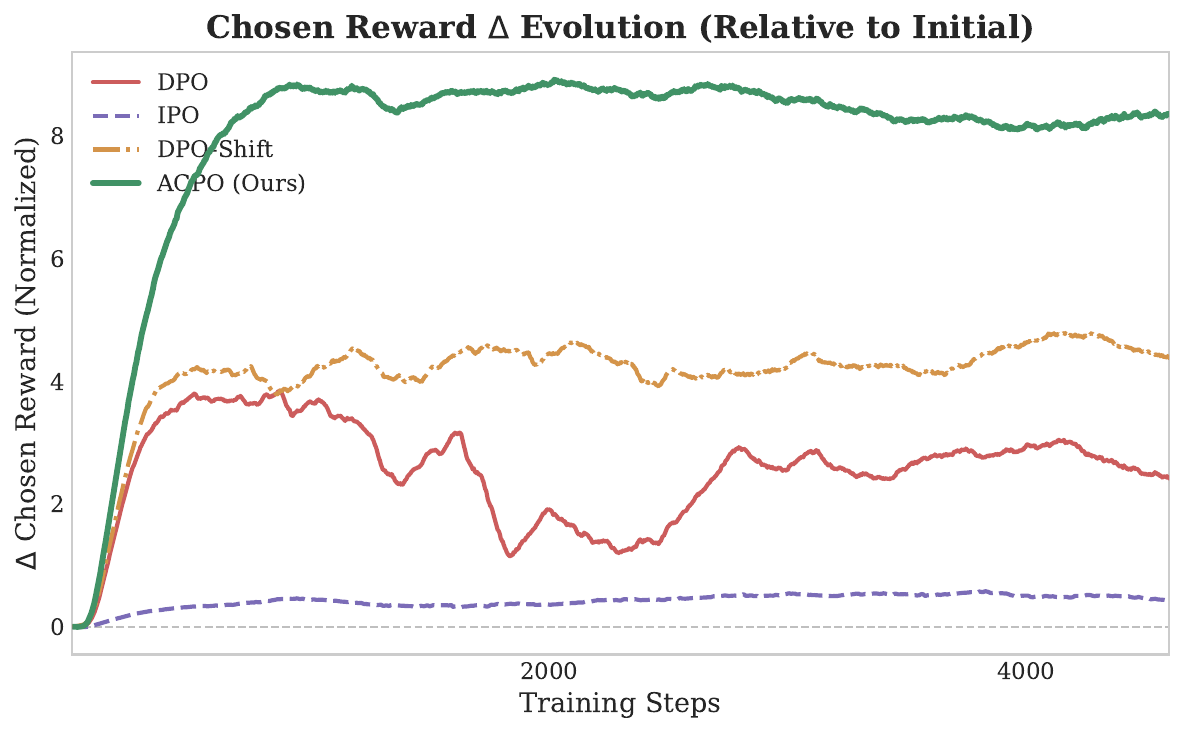}
        \caption{Chosen Reward Gain (\(\Delta r(y_w)\))}
        \label{fig:cross_method_a}
    \end{subfigure}
    \hfill
    \begin{subfigure}[b]{0.49\linewidth}
        \centering
        \includegraphics[width=\linewidth]{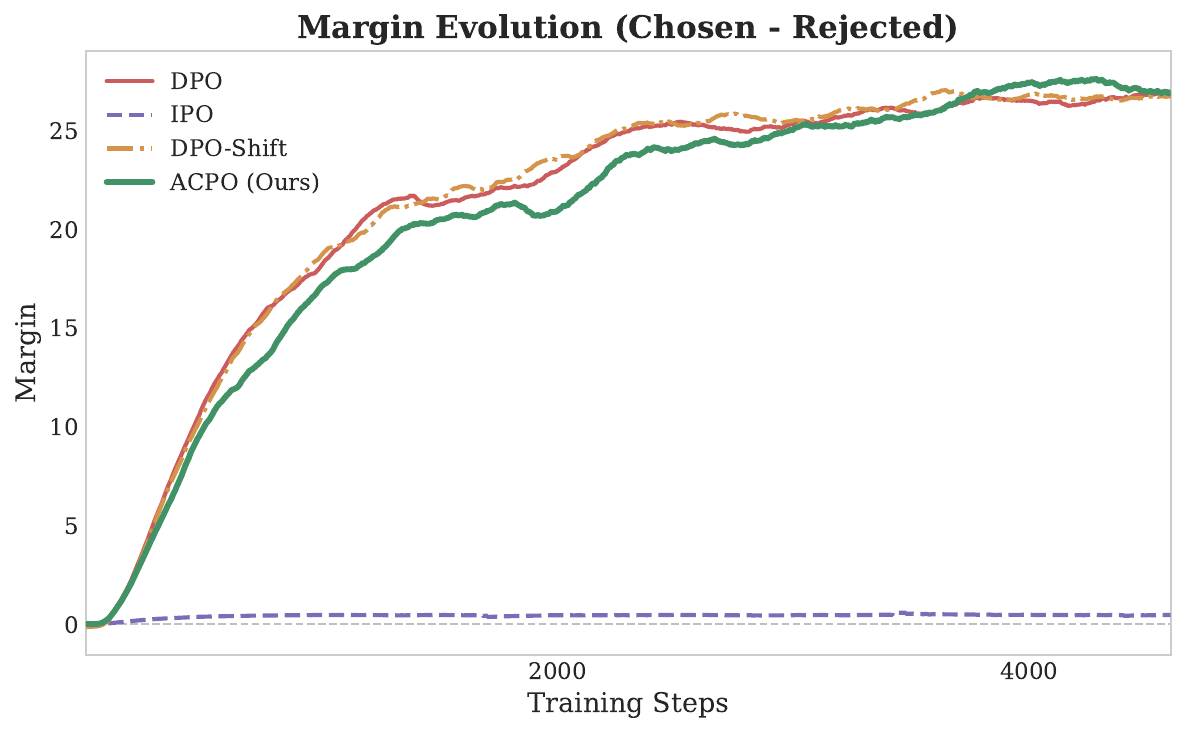}
        \caption{Margin Evolution (\(r(y_w) - r(y_l)\))}
        \label{fig:cross_method_b}
    \end{subfigure}
    \caption{\textbf{Cross-Method Comparison of Training Dynamics.} \textbf{(a)} Relative change in chosen reward from initial values. ACPO achieves the highest and most stable chosen reward gain ($\sim$+8.5), while standard DPO exhibits a pronounced drop after step 1000, consistent with Likelihood Displacement. \textbf{(b)} Margin evolution. All DPO-based variants converge to comparable margins ($\sim$27), indicating that ACPO improves chosen-reward preservation without sacrificing discriminative separation. SimPO is excluded due to its different reward scale (no reference model).}
    \label{fig:cross_method}
\end{figure}

\noindent\textbf{Cross-Method Reward Dynamics.}
To further contextualize these observations, we compare ACPO against DPO, DPO-Shift~\cite{yang2025dposhift}, and IPO~\cite{azar2024general} in Figure~\ref{fig:cross_method}. As shown in Figure~\ref{fig:cross_method_a}, ACPO achieves the highest relative chosen reward gain of $\Delta r(y_w) \approx +8.5$ and maintains a stable plateau throughout training. Standard DPO initially rises but suffers a marked decline after step 1000, consistent with Likelihood Displacement. DPO-Shift partially mitigates this decline yet still plateaus at roughly half the gain of ACPO. IPO exhibits negligible reward movement, consistent with its conservative regularization design.

Crucially, as shown in Figure~\ref{fig:cross_method_b}, all three DPO-based methods converge to nearly identical final margins of $\sim$27, with ACPO slightly overtaking the baselines after step 2500. This observation is central to our thesis: \textit{ACPO improves chosen-reward preservation without sacrificing the margin objective}. The decoupling of reward stability from margin optimization validates our asymmetric control mechanism as an effective solution to the Likelihood Displacement problem.

\FloatBarrier

\subsection{Ablation Studies}
\label{sec:ablation}
\begin{table}[!htbp]
\centering
\caption{\textbf{Ablation Studies on Core Components.} Removing either the asymmetric control or the dynamic length margin results in significant performance drops.}
\label{tab:ablation}
\begin{tabular}{l cc}
\toprule
\textbf{Model Variant} & \textbf{MMBench} & \textbf{POPE} \\
\midrule
\textbf{Full ACPO Framework} & \textbf{0.830} & \textbf{89.22} \\
\midrule
w/o Asymmetric Control ($\alpha=1$) & 0.830 & 86.89 \\
w/o Length Margin (\hlrev{$\tau_{\text{batch}} \to \beta$}) & 0.808 & 88.51 \\
\bottomrule
\end{tabular}
\end{table}

We dissect the ACPO objective to understand the contribution of its individual mechanisms. Results are evaluated on the POPE score \cite{li2023hallucination} and MMBench \cite{liu2024mmbench} overall score (Table~\ref{tab:ablation}):

\begin{itemize}
    \item \textbf{w/o Asymmetric Control ($\alpha = 1$):} When we force a symmetric update by disabling $\hat{\alpha}$, the objective formally degenerates to standard symmetric DPO. The model's visual anchoring degrades notably as a consequence, resulting in a significant drop in the POPE score (from 89.22 to 86.89). This highlights the absolute necessity of the unilateral anchor in preventing visual anchor collapse and avoiding the collateral suppression of valid tokens \cite{pal2024smaug}.
    \item \textbf{w/o Dynamic Length Margin (\hlrev{$\tau_{\text{batch}} \to \text{Static } \beta$}):} Replacing our complexity-aware target with a static scalar leads to performance degradation specifically in complex multimodal reasoning tasks, as evidenced by the drop in the MMBench score (from 0.830 to 0.808). Without length normalization, the gradient signal dilutes over long-context reasoning, exacerbating the well-documented length bias inherent in standard margin-based optimization \cite{meng2024simpo, park2024disentangling}, degrading the model's semantic density and general capabilities.
\end{itemize}

\subsection{Preference Evaluation and Attention Dynamics}
\label{sec:preference}
\begin{figure*}[t]
    \centering
    \includegraphics[width=0.95\textwidth]{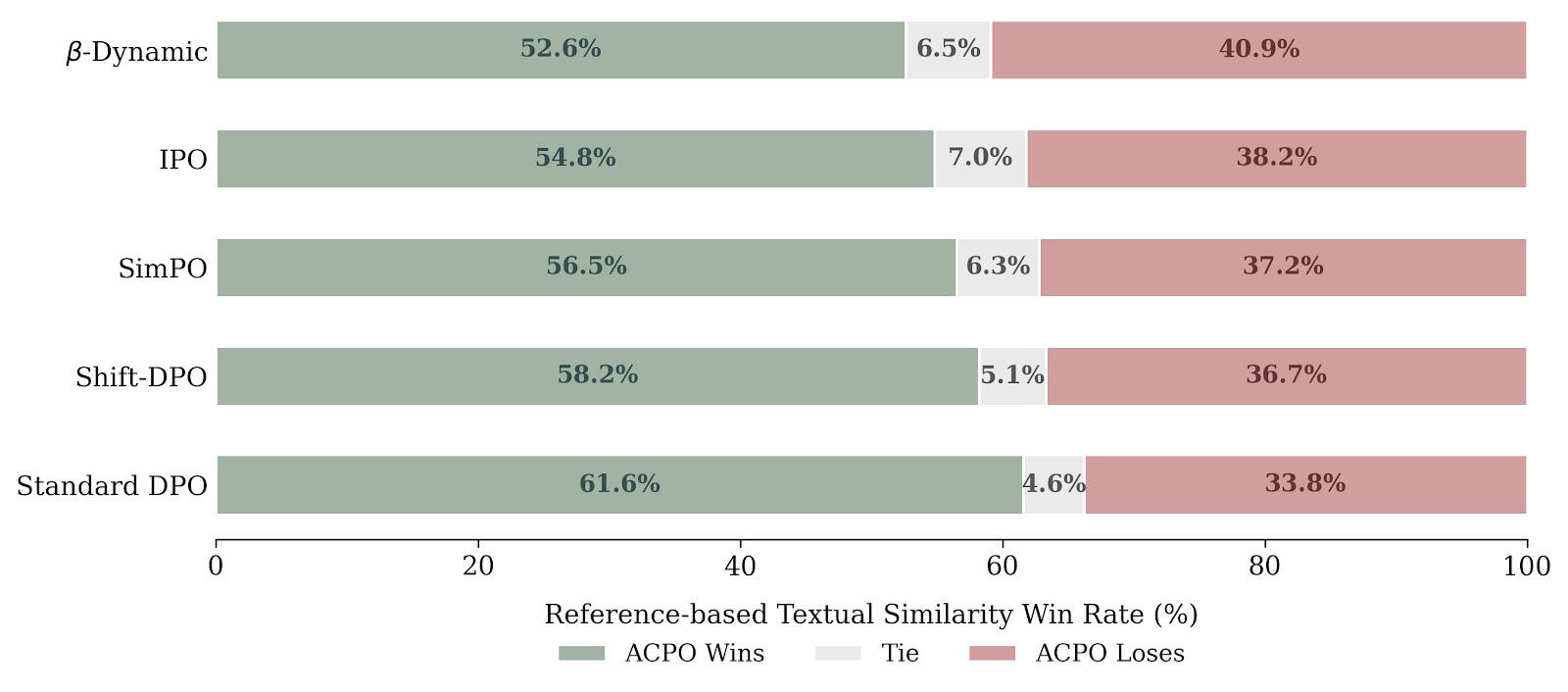}
    \caption{\textbf{Head-to-Head Preference Evaluation.} Pairwise win rates of ACPO against baseline methods, evaluated by Gemini. Using human-annotated \textit{chosen} responses as the gold standard reference, ACPO's outputs demonstrate significantly higher semantic alignment and fewer hallucinations compared to symmetric alignment methods.}
    \label{fig:win_rate}
\end{figure*}

\begin{figure*}[t]
    \centering
    \includegraphics[width=0.95\textwidth]{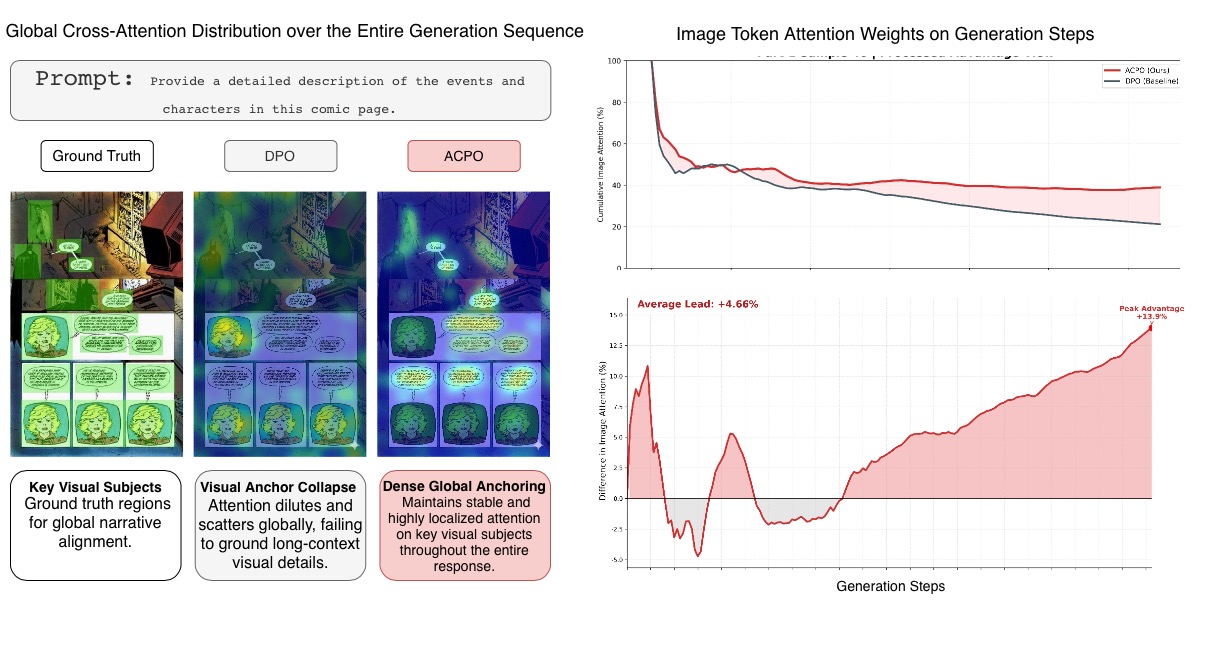}
    \caption{\textbf{Global Cross-Attention Distribution and Quantitative Tracking.} (Left) Heatmap visualizations of global attention averaged across all generated tokens. While visual anchor collapse is episodic, standard DPO is highly prone to it during long-context generation, causing attention to scatter to preceding text. ACPO maintains dense global anchoring on key visual subjects. (Right) Evolution of cumulative image token attention weights over generation steps, demonstrating that ACPO successfully arrests attention decay and maintains a significant advantage in visual grounding.}
    \label{fig:attention_dynamics}
\end{figure*}

To comprehensively validate whether the mitigation of Likelihood Displacement \cite{razin2025unintentional} translates into superior human-aligned generation, we conduct a head-to-head preference evaluation using Gemini as an impartial judge. Traditional metrics often fail to capture the nuanced subjective quality of multimodal reasoning \cite{chen2024mmstar}; thus, an LLM-as-a-Judge paradigm provides a rigorous assessment of alignment fidelity, aligning with recent advances in fine-grained visual evaluation \cite{yu2024rlhf}.

Using a hold-out test set, we employ the human-annotated \textit{chosen} responses ($y_w$) as the gold standard reference. We prompt Gemini to conduct blind pairwise comparisons between ACPO and baseline models (including Standard DPO, IPO, SimPO, $\beta$-DPO, and DPO-Shift). The judge is instructed to evaluate which model's output demonstrates a higher semantic alignment with the chosen anchor, exhibits stronger visual grounding, and produces fewer hallucinations. \textbf{To eliminate positional bias, each pair is evaluated twice by systematically swapping the presentation order. Furthermore, we verify that the average response lengths remain comparable across models, confirming that the evaluation is free from LLM verbosity bias.}

As illustrated in Figure~\ref{fig:win_rate}, ACPO achieves a commanding win rate against all baselines. Notably, when compared head-to-head against Standard DPO, ACPO is preferred in the vast majority of cases. \textbf{Even under high-temperature sampling where standard DPO frequently succumbs to language-prior hallucinations, ACPO maintains robust visual anchoring.} This decisively demonstrates that our asymmetric constraint directly improves the subjective quality of the model's responses by preventing the optimization process from drifting toward language priors.

To explicitly understand the underlying mechanism behind this subjective preference, we examine the global cross-attention distribution during inference. While visual anchor collapse is not strictly deterministic in every single generation, symmetric methods like DPO are highly susceptible to it, particularly as sequence length increases. Figure~\ref{fig:attention_dynamics} presents a representative case study of this pathological dynamic. As observed in the heatmap visualizations (Figure~\ref{fig:attention_dynamics}, left), standard DPO is prone to severe Visual Anchor Collapse; its attention diffuses globally and becomes anchored to preceding text rather than the required visual evidence. Conversely, ACPO's asymmetric constraint ensures Dense Global Anchoring, maintaining stable and highly localized attention on key visual subjects across multiple panels.

Furthermore, our quantitative tracking of cumulative attention over generation steps (Figure~\ref{fig:attention_dynamics}, right) corroborates this trend. While DPO's visual grounding often decays catastrophically as the sequence lengthens, ACPO effectively arrests this decay, sustaining an average lead of +4.66\% and a peak advantage of +13.9\% in visual attention retention. This conclusively demonstrates that our asymmetric objective mitigates the episodic vulnerability to language priors, directly improving response reliability and visual fidelity.

\section{Conclusion}
\label{sec:conclusion}

We proposed Asymmetric Constrained Preference Optimization (ACPO), a principled framework that addresses Likelihood Displacement in DPO-based alignment of Vision-Language Models. By introducing a Length-Adaptive Advantage Target and an asymmetrically scaled rejected penalty with stop-gradient, ACPO concentrates optimization pressure on suppressing incorrect outputs while preserving the chosen distribution's stability. Experiments on InternVL3 (14B and 8B) demonstrate state-of-the-art performance on hallucination benchmarks while driving concurrent improvements in general capabilities, and training dynamics confirm that ACPO effectively halts Visual Anchor Collapse.

\noindent\textbf{Limitations.} The current evaluation relies on a proprietary preference dataset. Validating ACPO on public preference corpora and extending it to multi-turn dialogue and online RL paradigms (e.g., GRPO) remain promising future directions.

%
%
\bibliographystyle{splncs04}
\bibliography{main}
\end{document}